\documentclass[times,authoryear]{elsarticle}

\usepackage{jasr}
\usepackage{framed,multirow}
\usepackage{float}

\usepackage{amssymb}
\usepackage{latexsym}

\usepackage{soul}

\usepackage[switch]{lineno}

\usepackage{url}
\usepackage{xcolor}
\definecolor{newcolor}{rgb}{.8,.349,.1}

\usepackage[citebordercolor=white]{hyperref}
\usepackage{graphicx}
\usepackage{subcaption}

\journal{Advances in Space Research}

\begin{document}

\verso{O'Keefe \textit{et al.}}

\begin{frontmatter}

\title{Practice Makes Perfect: A Study of Digital Twin Technology for Assembly and Problem-solving using Lunar Surface Telerobotics}

\author[1]{Xavier \snm{O'Keefe}}
\ead{Xavier.Okeefe@colorado.edu}
\author[1]{Katy \snm{McCutchan}}
\ead{Kathleen.Mccutchan@colorado.edu}
\author[1]{Alexis \snm{Muniz}}
\ead{Alexis.Muniz@colorado.edu}
\author[1]{Jack \snm{Burns}}
\ead{Jack.Burns@colorado.edu}
\author[2]{Daniel \snm{Szafir}}
\ead{daniel.szafir@cs.unc.edu}

\affiliation[1]{organization={Center for Astrophysics and Space Astronomy },
                addressline={University of Colorado Boulder},
                city={Boulder},
                state={CO 80309},
                country={USA}}

\affiliation[2]{organization={Department of Computer Science},
                addressline={University of North Carolina},
                city={Chapel Hill},
                state={NC 27599},
                country={USA}}


\begin{abstract}

Robotic systems that can traverse planetary or lunar surfaces to collect environmental data and perform physical manipulation tasks, such as assembling equipment or conducting mining operations, are envisioned to form the backbone of future human activities in space. However, the environmental conditions in which these robots, or ``rovers,'' operate present challenges towards achieving fully autonomous solutions, meaning that rover missions will require some degree of human teleoperation or supervision for the foreseeable future. As a result, human operators require training to successfully direct rovers and avoid costly errors or mission failures, as well as the ability to recover from any issues that arise on-the-fly during mission activities. While analog environments, such as JPL's Mars Yard, can help with such training by simulating surface environments in the real world, access to such resources may be rare and expensive. As an alternative or supplement to such physical analogs, we explore the design and evaluation of a virtual reality digital twin system to train human teleoperation of robotic rovers with mechanical arms for space mission activities. We conducted an experiment with 24 human operators to investigate how our digital twin system can support human teleoperation of rovers in both pre-mission training and in real-time problem solving in a mock lunar mission in which users directed a physical rover in the context of deploying dipole radio antennas. We found that operators who first trained with the digital twin showed a 28\% decrease in mission completion time, an 85\% decrease in unrecoverable errors, as well as improved mental markers, including decreased cognitive load and increased situation awareness.
\end{abstract}

\end{frontmatter}


\section{Introduction}
\label{sec1}

Space missions are increasingly dependent on robotic systems to achieve critical mission goals. Mobile manipulation robots in particular can offer a wide-range of mission support. For example, rovers like Curiosity and Perseverance have already demonstrated the value of sample data collection and analysis for scientific goals, with future systems envisioned to aid in tasks such as assembly (e.g., constructing habitats in advance of human missions or science facilities), infrastructure repair, equipment deployment, and mining. 

Such future rovers will play a critical role as the global space community has shifted its focus back to the Moon, with recent missions such as the United States' NASA Commercial Lunar Payload Services (CLPS) Intuitive Machine's Odysseus \citep{burns2021low}, India's Chandrayaan 3 \citep{rajasekhar2024comprehensive} and China's Chang'e 6 \citep{zeng2023landing}. Moon missions are expensive and challenging, and even small errors can have major consequences. For example, the Odysseus lander cost \$118 million and landed on its side due to stale telemetry, rendering some payloads unusable.

\begin{figure}[H]
  \centering
  \includegraphics[width=.7\linewidth]{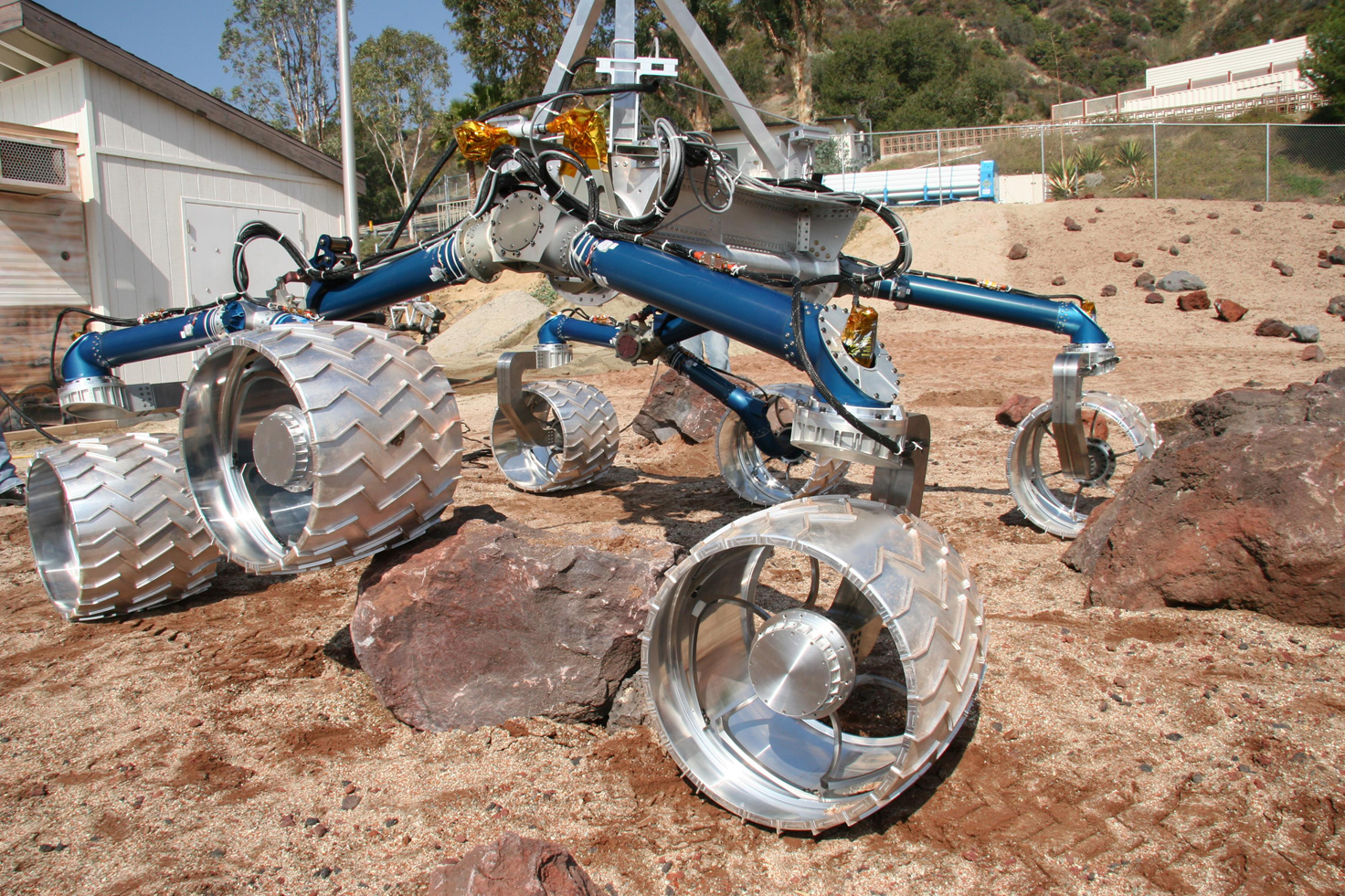}
  \caption{A mobility test of a rover in the NASA Mars Yard. The Mars Yard was created to physically simulate terrain on Mars and the Moon.}
  \label{fig:marsyard}
  \end{figure}

Due to the challenging nature of the lunar environment, future rover missions are not currently envisioned as fully autonomous, but will instead rely on some degree of human oversight and intervention. To achieve a seamless integration of human-robot systems and to mitigate the risks of high consequence errors for these operations, human operators will need to train comprehensively and have the ability to adapt to unforeseen issues and solve problems in real-time. 
How to best effect such training remains an open question as the operational environments of the Moon and Mars are not easily simulated physically on Earth. JPL's Mars Yard  (see Fig. ~\ref{fig:marsyard}), a state-of-the-art facility for large-scale physical simulations of other planetary environments, presents a compelling approach \citep{marsyard}. However, certain aspects, such as differences in gravity, cannot be easily simulated in the Mars Yard and access to its facilities is limited and costly. Moreover, in order to conduct testing and operator training in this facility, rovers must be transported to the Mars Yard, which can put hardware at risk, introducing unneeded cost. We are interested in exploring how to develop easier, cheaper, and more generalizable solutions for troubleshooting rovers and training operators to support upcoming space missions and keep pace with the growth of the commercial space economy, which may exceed the capacity of the Mars yard and other facilities like it.


Our approach is to develop a high-fidelity virtual environment and digital twin model that mimics an extraterrestrial setting, allowing for and more repeatable simulation of mission scenarios. Digital twin simulations also enable system operators to train on a less expensive yet accurate digital re-creation of the physical system and its environment, allowing operators to gain more applicable experience solving problems when compared to Earth-based physical simulations. In this research, we develop a physical rover and its digital twin to experimentally evaluate whether training and troubleshooting in a virtual environment enhances operator performance when controlling the physical rover during a simulated lunar mission. 

\section{Background and Related Work}
Robotics has become a key enabling tool for both autonomous and crewed space missions, where robots have served pivotal roles in missions across low Earth orbit (LEO), the Moon, and Mars \citep{gao2017review}. In this work, we focus on planetary rovers, robots that can operate on a planetary or lunar surface to collect environmental data, assemble equipment and infrastructure, and perform other physical manipulation tasks (e.g., mining). While substantial advancements in robotics technologies are continuing to push the envelope of autonomous robot capabilities, rovers and other systems still require substantial human oversight and intervention due to the complex and often uncertain nature of space mission environments and tasks \citep{nesnas2021autonomy}. Thus, it is essential that human team members are able to rapidly diagnose, respond to, and recover from any challenges that may arise during robot operations.

While this process is not yet fully understood, the Robot Failure Human Information Processing (RF-HIP) model has been proposed as a way to describe the complex interplay of human cognitive, emotional, and behavioral responses to robot failures \citep{honig2018understanding}. This model emphasizes three major aspects: how robots and associated systems may communicate failures, how humans perceive and comprehend failures, and how human-robot teams may solve failures. For example, research has explored how robots might leverage non-verbal cues, such as gaze behaviors, to communicate failures \citep{shiomi2013design} and how emotional responses, such as frustration in response to failures, can influence the effectiveness of human-robot collaborations \citep{weidemann2021role}. Likewise, failures are closely linked to human trust in robots \citep{law2021trust}, motivating recent research on robot proficiency self-assessment \citep{conlon2022m, conlon2024survey}.

Building on this body of research, we explore the development and evaluation of an interface for rover troubleshooting that leverages a \textit{digital twin}, a virtual representation of the robot and its environment. Digital twins, which add to the success of modern virtual reality technologies, have become transformative tools across a variety of disciplines by creating precise virtual representations of physical objects, processes, and systems. By leveraging a combination of pre-constructed models and real-time data, these digital replicas can enable predictive modeling, optimization, and simulation. For example, in manufacturing, digital twins can facilitate process automation, real-time monitoring, and predictive maintenance, leading to reduced downtime and increased efficiency \citep{lu2020digital}, while in emergency response, digital twins have been used to improve safety by helping humans monitor the evacuation status of a building during a fire \citep{ding2023intelligent} or enhance communication between offsite and onsite team-members \citep{walker2024cyber}. These technologies are being applied in urban planning, healthcare, and other parts of ``Industry 4.0'' to improve human decision-making by integrating multiple sources of data and enabling scenario testing without risking physical assets \citep{leng2021digital}. 

While virtual reality digital twin technology has become popular for terrestrial human activities, such as assembly pre-training in ground factories, it has yet to be widely adopted in the context of robotics, particularly space robotics \citep{mihai2022digital}. This context presents several new challenges, such as how to design and implement digital twins for platforms that are inherently mobile, rather than static, and operate in previously unseen or only partially known environments, how to capture and represent changing environmental surroundings in virtual simulations, how to develop accurate models and train machine learning algorithms for digital twin simulations given the relative sparsity of data surrounding space operation environments, and how to achieve necessary real-time data acquisition and communication using potentially low-bandwidth channels and low-power processors. Nevertheless, recent research is starting to explore how we might adopt aspects of virtual reality and digital twins for terrestrial robots as promising and intuitive mechanisms to aid users in understanding current and potential future robot states and actions \citep{walker2023virtual}. For example, a digital twin of a robotic arm has been developed as a training method for AI systems, allowing for training to occur without incurring wear on the real system \citep{matulis2021robot}. In this work, we present the first, to our knowledge, investigation of the use of a virtual reality digital twin in the context of operational training and troubleshooting for a lunar rover assembly mission.

Our mission context is inspired by an envisioned future telerobotic deployment of distributed arrays of surface sensors on the lunar far side to address multiple science goals recommended by U.S. National Academy Decadal Surveys. One such example is the NASA-funded concept called FARSIDE (Farside Array for Radio Science Investigations of the Dark ages and Exoplanets) \citep{burns2019nasa}. FARSIDE is a tethered array of thin-wire radio dipole antennas operating at low radio frequencies ($\approx$0.2-40 MHz) that is effectively impossible from the Earth, due to anthropogenic noise and a refractive and opaque ionosphere. The array consists of 256 dipole antennas distributed over a 10 km diameter in a four-arm spiral pattern. Another NASA-funded concept called FarView \citep{polidan2024farview} consists of 100,000 dipole antennas, constructed using in-situ manufacturing of dipoles from lunar regolith aluminum, with sufficient sensitivity to probe into the unexplored Dark Ages of the early Universe (before and during the formation of the first stars). The Astro2020 Decadal Survey's Cosmology Panel \citep{national2021decadal} stated that redshifted ``21 cm line intensity mapping of the Dark Ages [is] both the discovery area for the next decade and [is] the likely future technique for measuring the initial conditions of the universe in the decades to follow.'' The 21-cm wavelength radio radiation arises from neutral hydrogen that fills the Universe during this unexplored epoch, stretched via the expansion of spacetime to an observed wavelength of tens of meters (tens of MHz) \citep{loeb2012first}. The power spectrum of spatial fluctuations during the cosmic Dark Ages characterizes the amplitude of the variations as a function of spatial scale. During this time, the 21-cm line traces the cosmic density field with most modes in the linear regime, allowing a straightforward interpretation of the measurement in terms of the fundamental parameters of our Universe \citep{lewis200721}. The lack of luminous astrophysical sources makes the Dark Ages signal a clean and powerful cosmological probe and renders the 21-cm line the only electromagnetic observable signal from this era. Teleoperated (from Earth or the Moon) rovers with mechanical arms are key components for the construction of arrays that will open this new window to the Universe.

The first radio observations from the Moon recently began with the robotic commercial lunar lander, Intuitive Machines’ Odysseus,  touching down near the Malapert A crater in the South Pole region in February 2024. This mission was part of NASA’s CLPS program. The ROLSES (Radiowave Observations at the Lunar Surface of the photoElectron Sheath) radio science telescope was one of the NASA payloads on this lander \citep{burns2021low, hibbard2025results}. An upgraded ROLSES-2 on another CLPS lander is scheduled to situate on the lunar near side later in the decade. In addition, the CLPS Firefly Aerospace lander will deploy LuSEE-Night, NASA’s first radio telescope on the lunar far side, in 2026 with the goal of making trailblazer observations in the frequency band corresponding to the global redshifted 21-cm signal for the Dark Ages \citep{bale2023lusee}. These first radio telescopes will prepare us for the far side radio arrays expected in the 2030s.

\section{{System Design and Implementation}}
\begin{figure}[H]
\centering
\begin{subfigure}{.5\textwidth}
  \centering
  \captionsetup{width=.9\linewidth}

  \includegraphics[width=.9\linewidth]{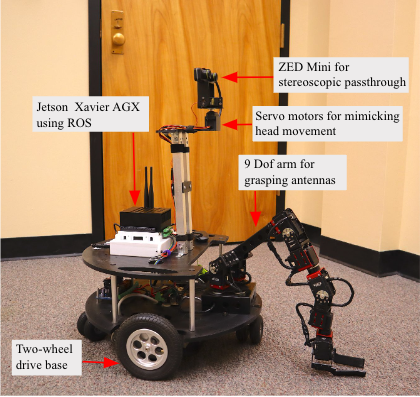}
  \caption{The physical Armstrong Rover in the Experiment room is shown above. Armstrong's arm, computer, cameras, and wheels are all mounted on the chassis.}
  \label{fig:physical}
\end{subfigure}%
\begin{subfigure}{.5\textwidth}
  \captionsetup{width=.9\linewidth}

  \centering
  \includegraphics[width=.9\linewidth]{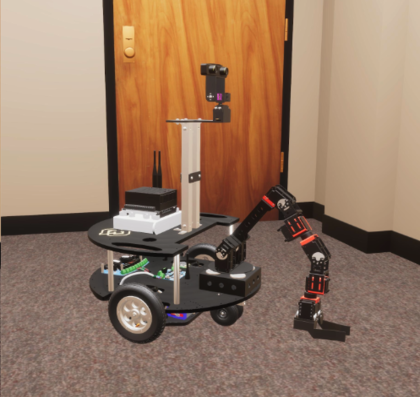}
  \caption{Armstrong's digital twin in the Unity environment is shown above. Note how similar Armstrong and the digital twin appear.}
  \label{fig:dt}
\end{subfigure}
\caption{A comparison of the physical Armstrong Rover and its digital twin in the Unity environment.}
\label{fig:armstrong_comparison}
\end{figure}

\begin{figure}[H]
  \centering
  \includegraphics[width=.9\linewidth]{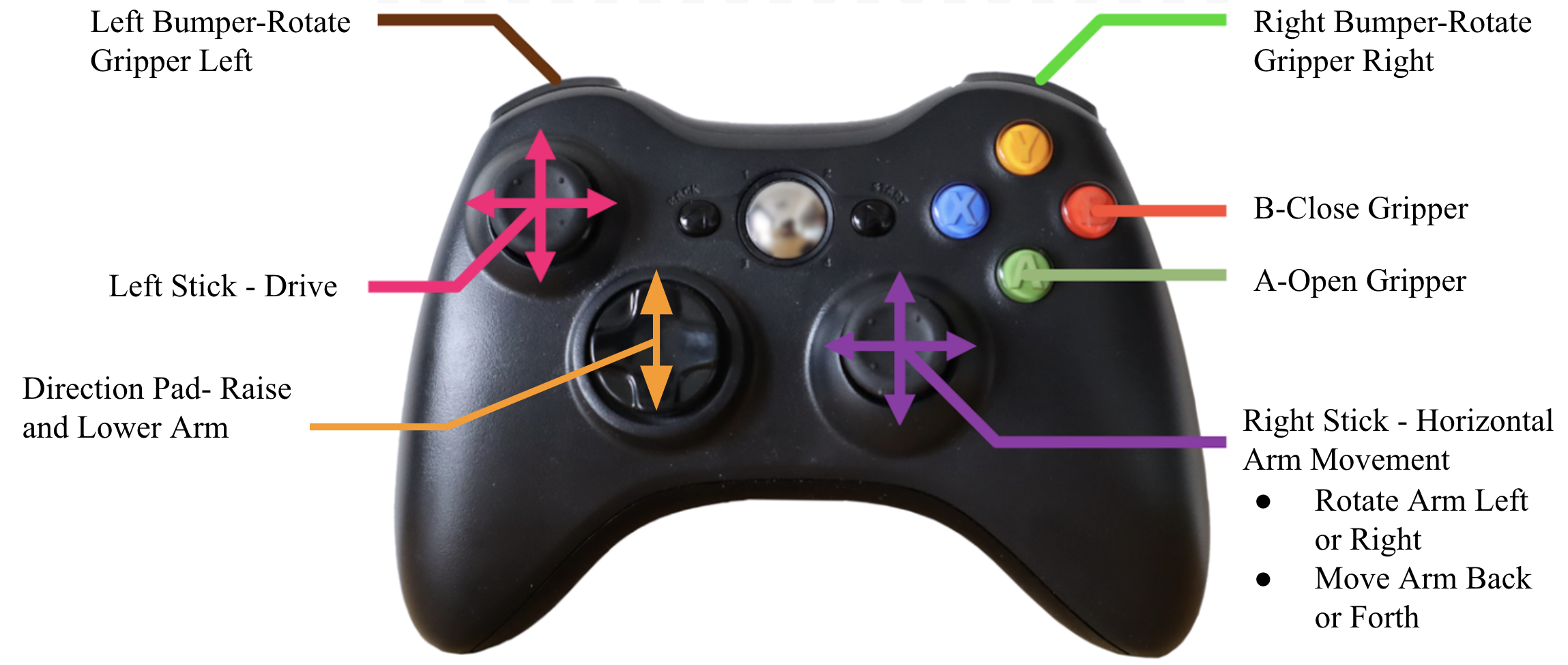}
  \caption{Users control both Armstrong and Armstrong's digital twin using an Xbox gaming controller. Armstrong's controls are based on similar systems in video games to reduce the learning curve for users.}
  \label{fig:controls}
  \end{figure}

 We designed and constructed a rover and mechanical arm system, dubbed ``Armstrong'' shown in Fig. \ref{fig:physical}, to provide a simple physical platform that was easily replicated in a virtual environment. We constructed Armstrong and its digital twin to investigate if training in a simulated environment with the digital twin will improve a user's ability to complete a task with the physical Armstrong rover. Armstrong was created as a testbed to perform this human factors research, and is not intended nor suitable as a flight model. Flight-ready rovers have complex dynamics and systems, and as a result are more difficult to model. Armstrong was deliberately designed with simplicity in mind, significantly reducing development time and enabling earlier implementation of the experiment.
 Armstrong is operated with an intuitive control layout based on the Xbox One gaming controller, as seen in Fig. \ref{fig:controls}.

\subsection{Robotic Platform}
 Armstrong has two drive wheels, allowing the user to drive forward and backwards, as well as pivot about the rover's vertical axis. See Fig. \ref{fig:physical} for Armstrong's Layout. Armstrong also has a 5 Degree of Freedom (DoF) arm with a gripper attachment. The arm can move side to side, up and down, and pivot about its mounting point on the chassis. The gripper can rotate about its attachment to the arm, as well as actuate its pincers. The drive and arm system are both controlled with an Xbox one controller. The button mapping of this controller was designed to replicate common mappings in popular video games to reduce the system's barrier to entry (see Fig. \ref{fig:build}). Armstrong's "brain" is a Nvidia Jetson Xavier AGX, a powerful computer often used in robotics. This computer runs the Robot Operating System (ROS) \citep{ros}, a popular open-source framework for robotics development that provides many convenient features out of the box. The Jetson is connected via WiFi to another computer running a first generation Oculus Quest through Unity, which streams video and head attitude data back to the rover. Unity is a popular game engine with many tools for VR applications. Unity's attitude data controls a pan-tilt servo system, allowing the user to "explore" Armstrong's environment in a natural way. The pan-tilt servos move a ZED mini stereoscopic camera, which streams VR-ready footage through the ZED-ROS interface. The integration of the pan-tilt servo system with the stereoscopic video from the ZED mini creates an immersive and natural viewing experience.  \\

 \indent
    \subsection{Digital Twin}
\begin{figure}[H]
  \centering
  \includegraphics[width=.9\linewidth]{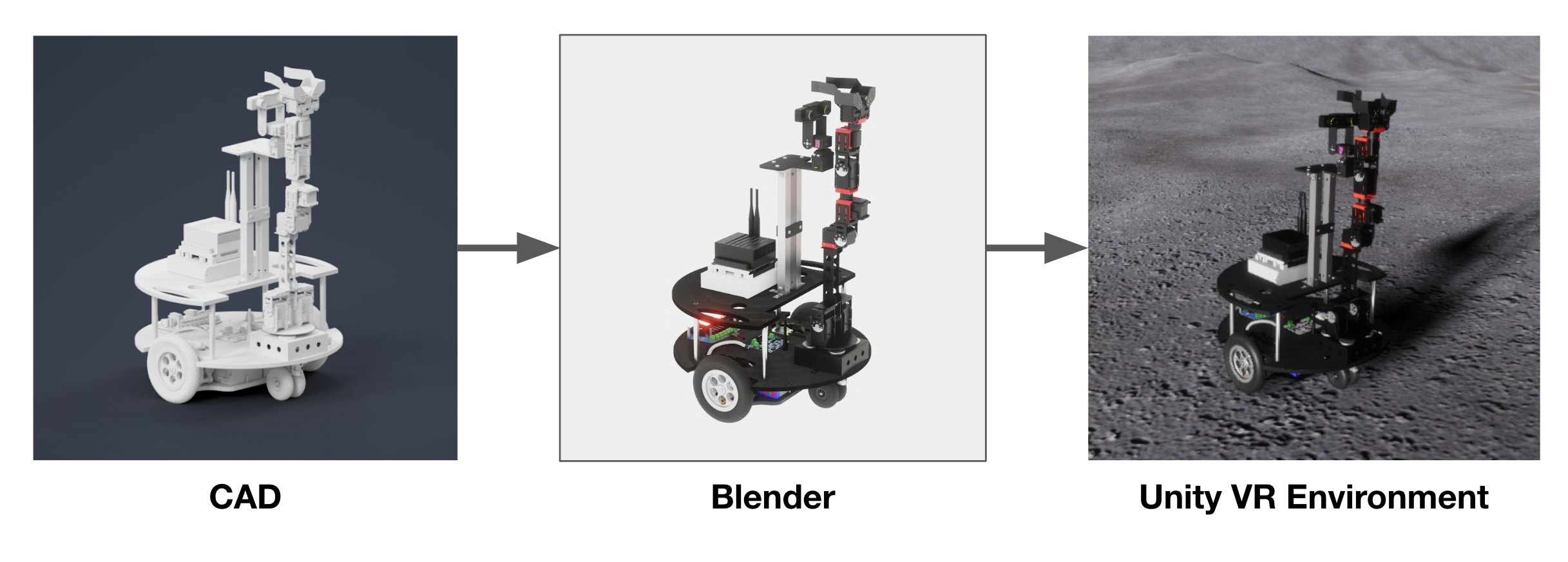}
  \caption{After the physical Armstrong rover was built, the team created a digital model using Computer Aided Design (CAD). Then, Armstrong's graphics were created in Blender, an Open-Source 3D graphical software. The CAD and graphics models were combined in Unity to make Armstrong's digital twin. }
  \label{fig:build}
  \end{figure}

 The virtual "digital twin" rover was designed to match the physical rover as closely as possible. See Fig. \ref{fig:build}. Each component of the physical robot and test environment were scanned and uploaded into Blender \citep{blender}, a CAD software used to modify 3-D models. We then carefully added texture and colors to each part, ensuring that each component was properly sized and looked nearly identical to its physical counterpart. 
 Each of these textured and colored components was assembled into the virtual arm model, with careful attention to maintaining accurate spacing and proportionality between elements. Once the virtual model was completely rendered and assembled in Blender we imported it into Unity \citep{haas2014history}, a popular game development software.
 
 With the digital twin digitized and loaded into Unity, we next focused on replicating the physical robot's movement characteristics. The team carefully measured the movement speed of each joint in the robots arm with a Python script, and verified these measurements with a human-operated stopwatch. The range of motion limits for each joint were also recorded. The robot's driving characteristics were measured over several simple tasks, which were timed and then averaged across all samples. After gathering and analyzing the data, we calibrated each arm joint's movement speed in Unity to match its physical analog. Configuring the robot's drive system required tweaking environmental variables (such as friction) in Unity until the robot was able to move through each task in the same manner and time as the physical rover. This was repeated until the physical and virtual rovers were able to complete the same series of benchmarks in roughly the same amount of time. These benchmarks measured two primary characteristics: movement speed and rotation speed. The movement speed benchmark measured the time required for Armstrong to move 1, 5, and 10 feet in a straight line in both forwards and reverse. The rotation benchmark compared the time it took Armstrong to rotate 90, 180, and 360 degrees. See tables ~\ref{tab:physical_virtual_times_2}, ~\ref{tab:physical_virtual_results}, and ~\ref{tab:physical_virtual} for the final results. All of these models and measurements were combined into a Unity program, forming the "virtual" Armstrong. The team conducted pilot studies early on in the experiment to solicit feedback from test operators on any differences between the physical and virtual rover. The virtual rover was then modified with any necessary changes.
\begin{table}[H]
\centering
\begin{tabular}{|c|c|c|c|c|c|c|c|}
\hline
\textbf{Distance (ft)} & \multicolumn{3}{c|}{\textbf{Physical Time (s)}} & \multicolumn{3}{c|}{\textbf{Virtual Time (s)}} \\ \hline
 & Trial 1 & Trial 2 & Trial 3 & Trial 1 & Trial 2 & Trial 3 \\ \hline
1   & 1.93 & 2.05 & 2.20 & 2.7 & 2.75 & 2.23 \\ \hline
5   & 9.75 & 9.78 & 10.07 & 9.85 & 9.29 & 10.53 \\ \hline
10  & 19.87 & 18.50 & 19.20 & 19.3 & 19.1 & 19.9 \\ \hline
\end{tabular}
\caption{Comparison of Physical and Virtual Drive Time in Forward Direction}
\label{tab:physical_virtual_times_2}
\end{table}

\begin{table}[H]
\centering
\begin{tabular}{|c|c|c|c|c|c|c|c|}
\hline
\textbf{Distance (ft)} & \multicolumn{3}{c|}{\textbf{Physical Time (s)}} & \multicolumn{3}{c|}{\textbf{Virtual Time (s)}} \\ \hline
 & Trial 1 & Trial 2 & Trial 3 & Trial 1 & Trial 2 & Trial 3 \\ \hline
1   & 2.18 & 2.20 & 2.57 & 2.56 & 2.25 & 2.61 \\ \hline
5   & 9.62 & 10.49 & 9.25 & 8.40 & 9.87 & 9.39 \\ \hline
10  & 20.74 & 18.30 & 18.38 & 19.22 & 18.80 & 19.52 \\ \hline
\end{tabular}
\caption{Comparison of Physical and Virtual Reverse Drive Time}
\label{tab:physical_virtual_results}
\end{table}

\begin{table}[H]
\centering
\begin{tabular}{|c|c|c|c|c|c|c|c|}
\hline
\textbf{Rotation (Degrees)} & \multicolumn{3}{c|}{\textbf{Physical Time (s)}} & \multicolumn{3}{c|}{\textbf{Virtual Time (s)}} \\ \hline
 & Trial 1 & Trial 2 & Trial 3 & Trial 1 & Trial 2 & Trial 3 \\ \hline
90  & 2.18 & 2.00 & 1.92 & 2.08 & 2.53 & 2.10 \\ \hline
180 & 3.70 & 3.75 & 3.87 & 4.12 & 4.58 & 3.70 \\ \hline
360 & 7.60 & 7.70 & 7.63 & 6.02 & 6.45 & 6.89 \\ \hline
\end{tabular}
\caption{Comparison of Physical and Virtual Rover Rotation Times}
\label{tab:physical_virtual}
\end{table}

\section{{Experimental Design}}
This study aims to explore the feasibility of low-latency telerobotic assembly of radio antennas in an array in the lab while also using a digital twin for problem-solving. While we did not simulate latency, such as that from the Earth to the Moon, this experiment serves as a first step in analyzing the feasibility of low-latency telerobotic assembly in space exploration missions on the Moon. This study also explores whether a digital twin can help with assembly failures and real-time troubleshooting. In addition, we aim to establish a baseline of user situation awareness (SA) \footnote{Situation Awareness is the term given to "the level of awareness that an individual has of a situation, an operators dynamic understanding of 'what is going on'" \cite{Salmon01072008} } and cognitive load (CL) \footnote{Cognitive load is "the amount of effort that is exerted or required while reasoning and thinking" \cite{interaction2016cognitive}} so that we can observe how the user's SA and CL vary from the baseline as other operation constraints are introduced in future experiments. 

\begin{figure}[H]
\centering
\begin{subfigure}{.45\textwidth}
  \centering
    \captionsetup{width=.9\linewidth}

  \includegraphics[width=.9\linewidth]{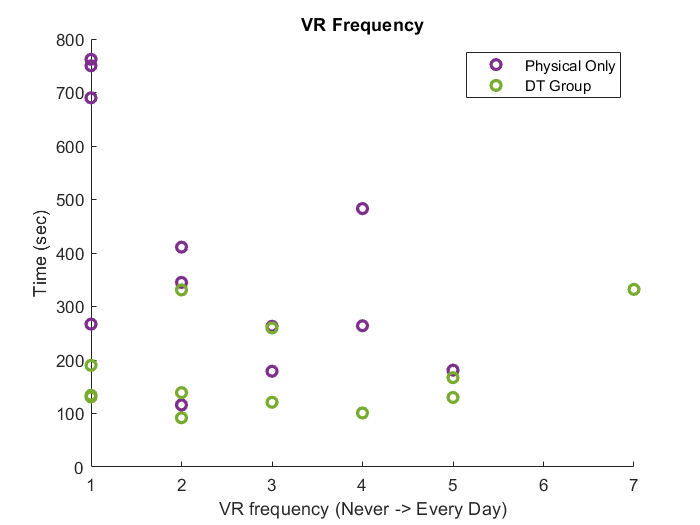}
  \caption{VR use frequency versus time. Physical only corresponds to Group A and DT (Digital Twin) corresponds to Group B. Previous experience with VR is not a serious bias among the participants.}
  \label{fig:test1}
\end{subfigure}%
\begin{subfigure}{.45\textwidth}
  \centering
    \captionsetup{width=.9\linewidth}

  \includegraphics[width=.9\linewidth]{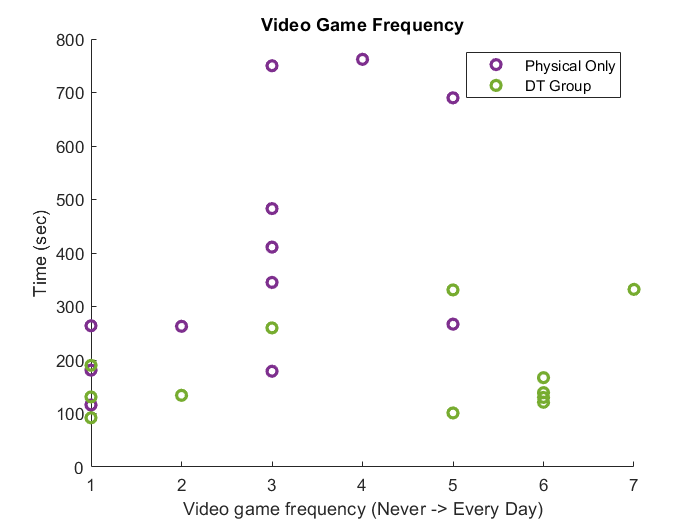}
  \caption{Video Game use frequency versus time. Physical only corresponds to Group A and DT (Digital Twin) corresponds to Group B. Previous use of video games does not appear to be a major factor in the experimental results.}
  \label{fig:test2}
\end{subfigure}
\caption{Comparative analysis of VR use frequency and Video Game use frequency over time among different groups.}
\label{fig:frequency_comparison}
\end{figure}

We hypothesize that Armstrong's digital twin operation will improve participants' performance on the physical Armstrong rover by having a faster completion time than participants who only complete the task on the physical rover. The experiment was designed as a 2x1 study in which subjects were randomly assigned a condition, and subjects of each group were compared to the performance of the other group. Group A was the group that drove the physical rover only for one run to provide a baseline, and Group B operated the digital twin rover once before operating the physical rover once. Group A was intentionally designed to \textit{not} operate the physical rover twice, as the experiment was conceived to evaluate the applicability to space scenarios. One cannot train in a fully realistic environment, so the mission is often the first time the operator is able to interact with the system in realistic conditions. Our study replicates this with group A, and compares their results to the trained operators of group B to show that a practice run with our digital twin platform effectively trains operators for a mission. In this experiment, participants were tasked with teleoperating the Armstrong rover using immersive virtual reality to align misaligned antennas. We recruited 24 individuals from the University of Colorado Boulder campus to participate in the study approved by the University's Human Subjects Institutional Review Board (IRB). Participants ranged in age from 18 to 60, with an average age of 24.3. There was variation in the frequency of participants who used gaming and/or VR consoles; however this did not impact the distribution of the final times, as shown in Fig. \ref{fig:test1} and Fig. \ref{fig:test2}. 

\subsection{Experimental Task}
\begin{figure}[H]
  \centering
  \includegraphics[width=.7\linewidth]{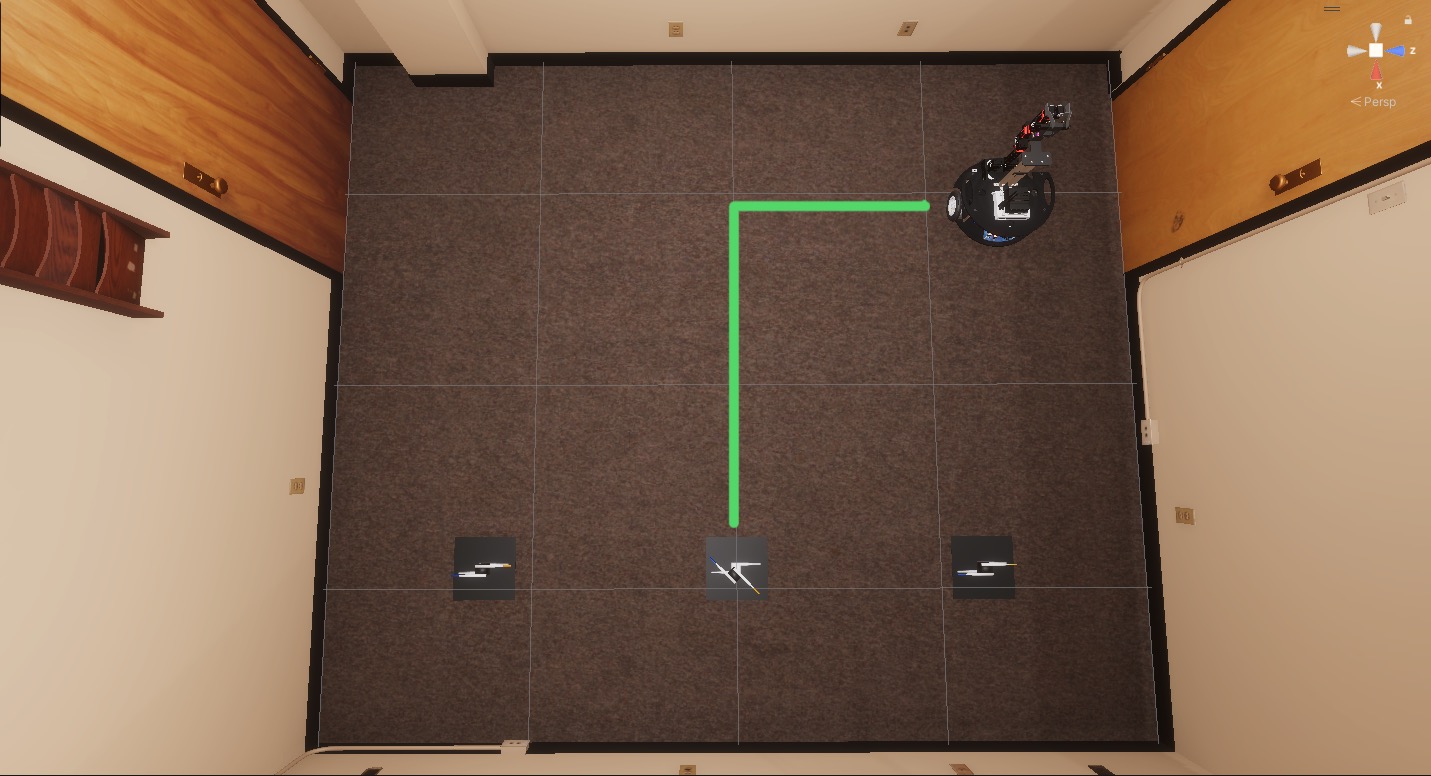}
  \caption{Overhead view of a trajectory participants could take to complete the task. Armstrong begins in the top right corner of the room facing the wall. Participants must turn Armstrong around and navigate to the misaligned antenna. }
  \label{fig:overhead}
  \end{figure}

This experiment involved participants teleoperating a small robot equipped with a robotic arm (Armstrong) in a manner analogous to operating a rover/arm on the Moon. Participants were tasked with aligning dipole radio antenna units using the telerobotic interface (see Fig.  \ref{fig:controls}). Participants wore a virtual reality (VR) headset and controlled the rover with the Xbox video game controller. Video feedback was sent through a ZED mini camera for both the physical and digital twin rovers to the headset, acting as the eyes of the robot. The ZED mini is a mixed-reality camera that captures stereoscopic video similar to human binocular vision, making it ideal for VR/AR applications. Half of the participants teleoperated only the physical robot (Group A), while in the other trials, participants first operated a digital twin in a virtual environment before controlling the Armstrong physical rover (Group B). The participants had to first drive the rover to an array of antennas. See Fig. \ref{fig:overhead} for an example trajectory. The participant then used Armstrong's cameras to identify the misaligned antenna and subsequently realign the antenna using Armstrong's gripper. To realign the antenna, participants must lower the arm, open the gripper's jaws, close the gripper around the antenna, either rotate Armstrong's wrist or reposition the rover itself, put the antenna down, and remove the gripper from the antenna. See the \href{https://www.youtube.com/watch?v=6YKsZBBNcwU&ab_channel=NetworkforExplorationandSpaceScience}{attached informational video} for an illustration of the experimental task in ~\ref{sec:video}.

The task was chosen to resemble a sample task for the FarView radio interferometer mission \citep{polidan2024farview}. It was designed to be easy enough for an inexperienced first-time operator to complete within a reasonable time, while also not being trivial. These metrics were evaluated in a series of pilot studies, in which operator performance on the task was found to fall within an acceptable threshold.

\subsection{Procedure}
This study used 24 subjects for the experiment, with each subject completing the experiment individually (i.e., multiple subjects will not run the experiment simultaneously). This study is a within-participants design. The subjects operated in two separate locations throughout the experiment. Group A operated in the Command Room. The Command Room is physically isolated from all robots, and subjects used the VR headset with video feedback and a video game controller to send commands to the robot (which is in the Operations Room). The subject was in the Command Room during the Training Phase and data collection. The training phase served as an opportunity for all subjects to become familiar with our system's control and video interface. The controls were explained, and participants were allowed 10 minutes to practice with the robot before trials begin. The assembly task was also presented during the Training Phase. Three main sub-tasks were required to be completed correctly. First, the rover was required to drive to the misaligned antenna. Next, the rover's gripper  was tasked with gripping the antenna. Finally, the antenna needed to be aligned with a place marker in the designated location. The antenna unit is complete after the antenna is aligned in the correct position. When operators inadvertently manipulated the antenna into an unrecoverable orientation, the timing was stopped and the antenna was reset to its original position. An unrecoverable position was defined as one in which the antenna was not resting on its bottom face. After resetting the antenna, participants were allowed to resume their trial. The trial was marked as a failure if and only if the participant was unable to sufficiently align the antenna within the allotted 15 minute time window. A video recording of each participant from both first and third person angles were used to determine antenna alignment times and identify any alignment errors.

After completing the Group A task (either by finishing the alignment task or by the time running out), the subject was asked to complete a series of surveys, outlined in greater detail in ~\ref{subsubsec:Subjective_measures}.

In Group B, the subjects were instructed to begin their assembly tasks while sending commands from the Command Room. Group B completed two tasks. Group B performed the same physical task but also performed a virtual version of the physical task first. They used the virtual headset and controller connected to a computer to control a virtual version of the rover. The assembly task during Group B was identical to the task in Group A, except it included using the virtual environment as a practice environment. After completion of Group B (either by finishing all assembly tasks or by the time running out), the subject once again completed the same surveys as Group A but also answered questions relating to the virtual environment. The results of Groups A and B (both the recorded video and the surveys) were compared to determine how human cognitive load changes when operating via low-latency telerobotics instead of operating with a direct physical presence in the operating area.

\subsection{Measures and Analysis}
\label{subsec:measures_analysis}
{
\subsubsection{Objective Measures}
Several objective measures were used to quantify a participant's performance. The first of these measurements was time to completion, which was recorded manually by the experimenters. The next measurement was the number of "antenna flips", which was defined as the amount of times that the operator manipulated the antenna into an unrecoverable position. When the antenna was flipped into an unrecoverable position we stopped the timer, instructed the participant to move the rover away from the antenna platform, reset the antenna to its original position, and resumed time. We also measured the number of collisions with obstacles in the experiment room. The Student's t-tests were used to evaluate if there was a significant difference between the means of any two samples taken across both groups. T-tests results gain much more meaning when extra statistics are reported, which we have done in this paper. Our t-tests results have the form t(n) = t statistic, p = p-value, d = Cohen's d. These definitions are tabulated below, where the null hypothesis claims that the means of the populations of the two groups being studied are equal \citep{boslaugh2012statistics}.

\begin{center}
\begin{tabular}{|p{1.5cm}|p{11cm}|}
    \hline
    \textbf{Variable} & \textbf{Definition} \\ \hline
    n & Degrees of Freedom. Higher degrees of freedom, usually associated with larger sample sizes, lead to more precise estimates and more reliable confidence intervals. \\ \hline
    t statistic & Measures how far the sample mean is from the hypothesized population mean in terms of standard errors. A larger t statistic indicates a larger difference relative to the variation in the data. Values over 2 are considered acceptable, and indicate a rejection of the null hypothesis, so long as the p-value is also significant.  \\ \hline
    p-value & Represents the probability of obtaining the observed data, or something more extreme, under the assumption that the null hypothesis is true. A p-value below 0.05 is conventionally considered significant, indicating potential rejection of the null hypothesis.\\ \hline
    Cohen's d & Measures the effect size, indicating the magnitude of the difference between two means relative to the standard deviation. A d value of 0.8 or above is considered to represent a large effect size, indicating that the treatment was effective.\\ \hline
\end{tabular}
\end{center}

\subsubsection{Subjective Measures}
\label{subsubsec:Subjective_measures}
We used several subjective measures to quantify the mental state of our participants and the usability of our system. 

\begin{itemize}
    \item The NASA Task Load Index (TLX) survey is used to gauge subjective workload among participants using six key attributes: mental demand, physical demand, temporal demand, performance, effort, and frustration, each evaluated on a scale from low to high \citep{hart1988development}. The results from this survey were used to quantify the mental, physical, and temporal demands of the experiment, as well as the perceived success, effort, and frustration of the participant. These results were also used to verify that our operators were not mentally over-taxed in any of the task components. See ~\ref{subsec:TLX} for the full list of questions.
    
    \item The System Usability Scale (SUS) is a Likert scale survey that is widely used to gauge a user's experience with a system. SUS results are considered to be "quick and dirty" \citep{brooke1996sus}, but suitable for general evaluations on the usability of a system. Participants answered a range of questions, and the results were placed into a standard formula to calculate a system usability score. The results of this survey were used to verify that our systems (a) performed consistently and (b) did not hinder the operators in any way. See ~\ref{subsec:sus} for a list of these questions.
    \item The Affect Grid is a self-reporting scale that is designed to measure how the participant is emotionally reacting to the experiment \citep{affectgridrussell}. The Affect Grid uses two different dimensions to measure how the participant is feeling. One dimension measures positive or negative emotions and the other dimension samples alertness vs. sedation. Throughout the experiment, participants rated how they were feeling on both of these aspects. These results were used to monitor participants transient mood throughout the experiment, and to ensure that participants were not overly stressed or anxious. See ~\ref{sec:affect_grid} for an image of the Affect Grid administered in this experiment.
    \item Questions developed by the team. We asked several questions to gauge specific attributes of our experiment, such as the continuity between digital and physical experiences, and the ease of manipulating objects. We also surveyed users on their  past experiences with VR and video games, as well as their age and gender. The results from these questions were used to identify trends related to any of these metrics, and to evaluate the effectiveness of our study throughout a wider population. See ~\ref{subsec:researcher} and ~\ref{subsec:demographics} for a list of these questions.
\end{itemize}
}

\section{{Results}}
\subsection{Objective Results}
\begin{figure}[H]
\centering
\begin{subfigure}{.5\textwidth}
  \centering
      \captionsetup{width=.9\linewidth}

  \includegraphics[width=.9\linewidth]{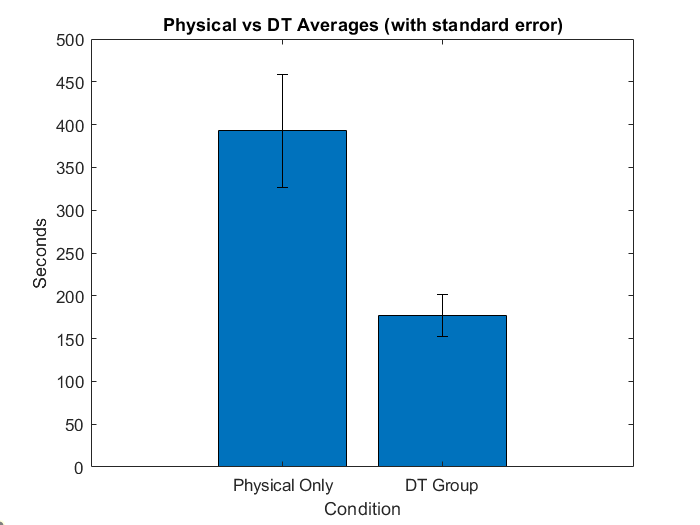}
  \caption{Histograms of average completion times for both groups, with standard error. Note the difference between the Physical Only group and the DT group.}
  \label{fig:time}
\end{subfigure}%
\begin{subfigure}{.5\textwidth}
  \centering
      \captionsetup{width=.9\linewidth}

  \includegraphics[width=.9\linewidth]{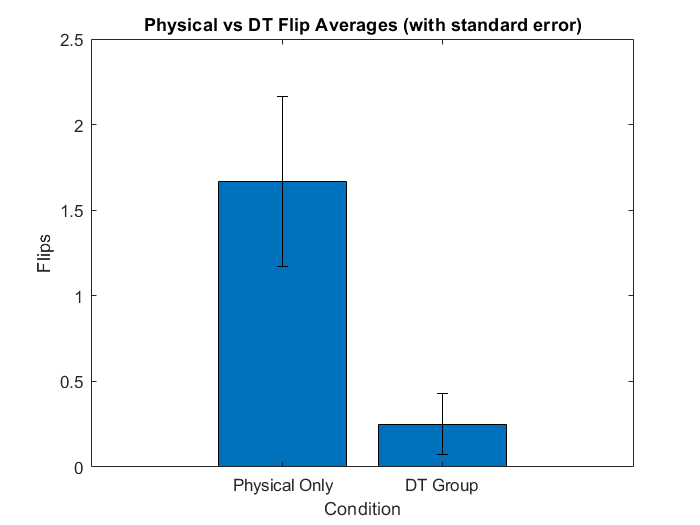}
  \caption{Histograms of antenna flip amounts for both groups, with standard error. Note that the Physical Only group had a much greater number of flips.}
  \label{fig:flips}
\end{subfigure}
\caption{Comparison of completion times and antenna flip amounts between groups with standard error.}
\label{fig:comparison}
\end{figure}

{
We analyzed task performance and cognitive loading to determine if our digital twin platform helped operators to complete tasks more effectively when compared to operators who only used the physical Armstrong rover/arm. We found a highly significant effect of our digital twin platform on time to completion, using the student t-tests with t(22) = 3.05, p = 0.006, d = 1.25, with the digital twin training platform decreasing time to completion by 28\%. See Fig. \ref{fig:time}. These t-tests results show a significant difference in means between samples, with a p-value that indicates a strong level of rejection of the null hypothesis, and a Cohen's d value indicating a large effect size.
\\
\indent 
We also found that the digital twin platform greatly reduced the amount of times that the operator "failed" the mission, which was determined by the number of times the antenna was manipulated into an unrecoverable position. We found that the digital twin platform reduced these instances over 85\%, again with the t-tests, with t(22) = 2.68, p = 0.014, and d = 1.09. See Fig. \ref{fig:flips}. Both of these t-tests results show a significant difference in means between samples, with a p-value that indicates a strong level of rejection of the null hypothesis, and a Cohen's d value indicating a large effect size.
}

\subsection{Subjective Results}
{
Several interesting results emerged when analyzing the responses to our administration of the NASA TLX survey. The first, shown in Fig. \ref{fig:timepressure}, was a 36\% decrease in the operator's perceived time pressure during the experiment. This shows that training with the digital twin not only reduced the time in which participants completed the task, but also reduced their stress levels during the trial. The second interesting result was a 4\% decrease in in mental demand between the two groups, shown in Fig. \ref{fig:mentaldemand}. We expected that the perceived mental demand would decline between the two groups, but our results suggest that both groups experienced roughly the same mental stress. This could be due to the particular interface with the technology - virtual reality is not yet commonplace, and participants may have had difficulty mastering it. It is worth noting both groups reported a medium mental demand. The third interesting result was 17\% decrease in frustration levels between the subjects, shown in Fig. \ref{fig:insecurity}. This corroborates the first result, further confirming that training with the digital twin reduces stress in operation, which will ultimately result in fewer errors.

Another interesting result from our surveys were the responses of each group to the question ``It was easy to grip the antenna'', with Group B reporting a 35\% easier time gripping the antenna, shown in Fig. \ref{fig:grip}. Gripping the antenna was the most challenging part of the task, so the substantial performance enhancement observed after digital twin training suggest promising implications for skill transfer in teleoperation interfaces. This result shows that training with the digital twin not only improves the operator's mental state, but also improves their comfort performing difficult tasks.
\begin{figure}[H]
  \captionof{figure}{Histograms showing subjective results}
    \begin{subfigure}[t]{.5\textwidth}
        \centering
            \captionsetup{width=.9\linewidth}

        \includegraphics[width=\linewidth]{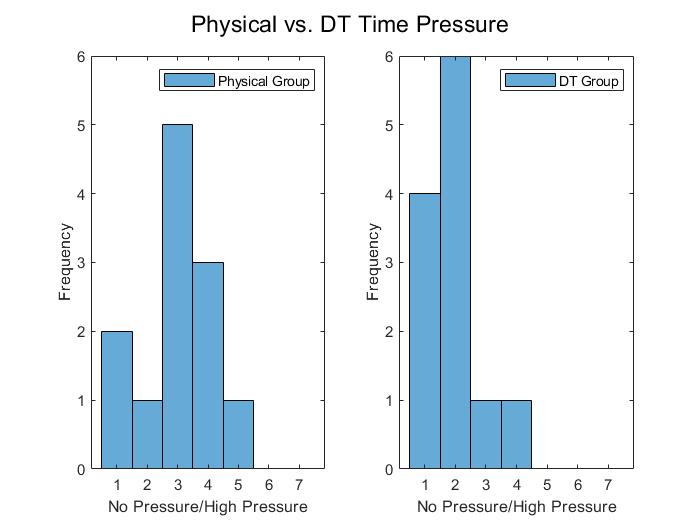}
        \caption{Self reported time pressure felt by users during physical and digital twin tasks. Users in the DT group felt significantly less time pressure.}
        \label{fig:timepressure}
    \end{subfigure}
    \hfill
    \begin{subfigure}[t]{.5\textwidth}
        \centering
            \captionsetup{width=.9\linewidth}

        \includegraphics[width=\linewidth]{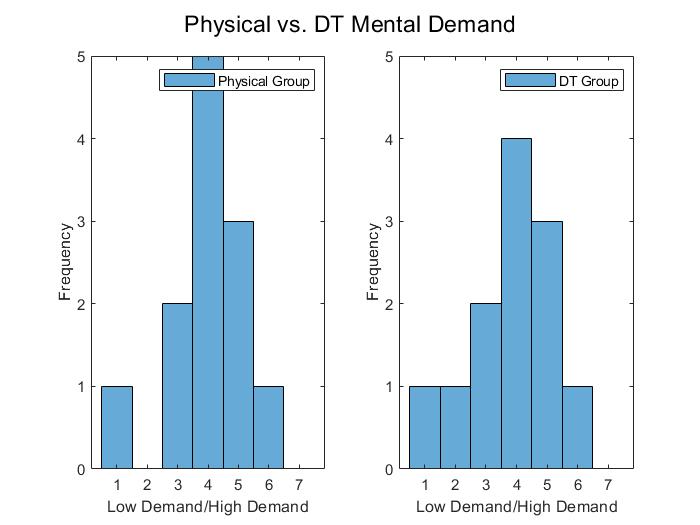}
        \caption{Self reported mental demand felt by users during physical and digital twin tasks. Note how similar the graphs are.}
        \label{fig:mentaldemand}
    \end{subfigure}

    \medskip

    \begin{subfigure}[t]{.5\textwidth}
        \centering
            \captionsetup{width=.9\linewidth}

        \includegraphics[width=\linewidth]{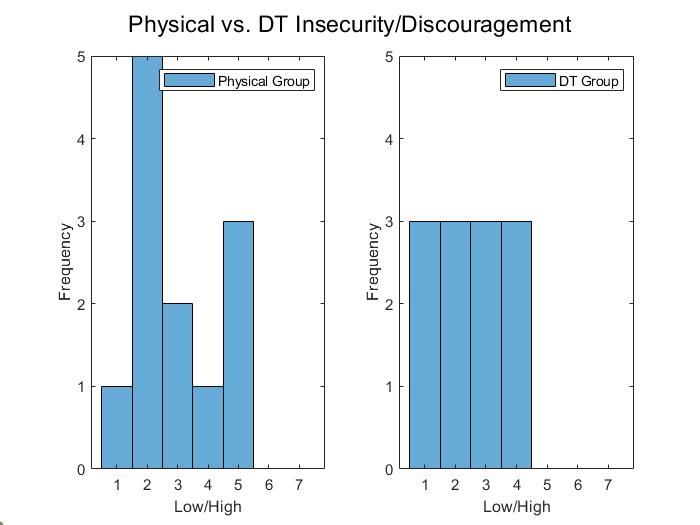}
        \caption{Self reported insecurity and discouragement felt by users during physical and digital twin tasks. Analyzing the means of these results showed lower stress levels in the DT group.}
        \label{fig:insecurity}
    \end{subfigure}
    \hfill
    \begin{subfigure}[t]{.5\textwidth}
        \centering
            \captionsetup{width=.9\linewidth}

        \includegraphics[width=\linewidth]{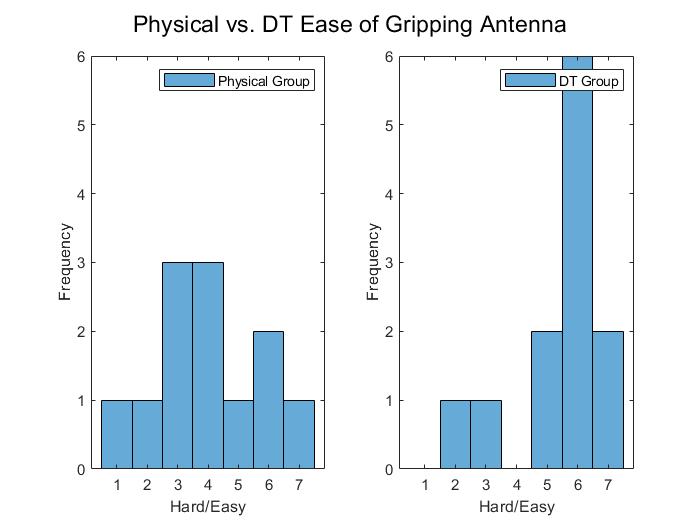}
        \caption{Self reported ease of gripping antenna by users during physical and digital twin tasks. The digital twin group found it much easier to grip the antenna.}
        \label{fig:grip}
    \end{subfigure}
\end{figure}
\indent

We also evaluated our system usability using the System Usability Scale (SUS). We did not find a significant difference, t(22) = 0.2075, p = 0.84, d = 0.09, between the digital twin platform (M = 71.45) and the Physical platform (M = 70). Both of these system usability scores qualify as "above average". This indicates that any differences in performance between the two groups are a result of training with the digital twin, rather than differences in interfaces.

Results from the Affect grids were also analyzed, and found to remain fairly consistent across participants. This shows that participants did not feel overly fatigued throughout the course of the experiment, and that the platform was not overly mentally demanding. This adds an additional layer of robustness to our results because it further verifies the similarity of the platforms by showing that operator experience remained nearly identical. These results indicate that the differences seen in operator performance between the two groups are in fact due to training with the digital twin, and not some difference in platform or usability. 
}
\section{{Discussion}}
{
As humanity shifts its focus back to the Moon, the need for training on mission-specific tasks will only increase. The exact features of the Moon's environment are impossible to physically replicate on Earth, which is why it is important to develop realistic simulations. Our experiment has shown that these simulations can provide helpful and applicable training and problem-solving to operators, without the need to be physically present in the target environment or endanger expensive equipment. Our system was proven to be usable via the System Usability Scale, showing that operators are comfortable training in a virtual environment. Both the digital and physical systems received similar SUS scores, suggesting that the significant objective results were due to the impact of training with the digital twin, rather than differences in operator experience with each system. Training with the digital twin platform also reduced the user's perceived cognitive load, decreasing time-pressure and frustration while operating. Digital twin training also improved operator's comfort while manipulating the antenna, which was easily the most difficult part of the task. Results from the Affect Grid show that participants moods were not impacted by the experiment or training, showing that our task was not too difficult, and that our system did not significantly tire the operators. All of these results combined show that training with the digital twin provides significant improvements to the operator's experience, and therefore performance.
\\
\indent
The results of the study also show that training with the digital twin platform significantly increased operator efficiency. Participants that trained on our digital twin platform showed dramatically reduced time to completion, as well as markedly increased accuracy in completing the task, as measured by the difference in unrecoverable antenna flips. We observed striking differences between the physical only and digital twin groups in these two areas, strongly supporting our hypothesis.
\\
\indent
Another area that was studied in the data was the suitability of our task for evaluating the digital twin platform. If the task was too simple, one would expect all operators to complete it with ease in a similar amount of time. If the task was too complex, we would expect several operators not to complete it in the allotted time. Neither of these effects are evident in the data - we saw large standard deviations in the time to completion for each data set, with no participants failing to complete the task in the given 15 minutes. These markers indicate that our task selection was sufficient to evaluate the platform.

\indent
These results are significant for the space community as a whole - they show that it is not only possible but also practical to train operators with a virtual simulation. This technology should reduce the human risks associated with a mission's success, and help more planetary surface missions succeed. This technology will also be useful in a much wider range of scenarios, from deep sea robotics to above-ground disaster situations and agriculture. Practically any operation in the world could benefit from a portable training method that costs less money, risks less equipment, and provides adequate training to operators. 
} 

\section{{Limitations}}
{

Although our work has shown promising results in regards to digital twin training and its applicability to upcoming lunar missions, it is not without its limitations. One of these limitations is that our mock antenna alignment is both very simple and not real - meaning that actual results on the Moon may be different than observed. In a real lunar mission the stakes are much higher, with increased levels of latency and increased uncertainty in the environment. The simple nature of our task may have also impacted participant performance, and more work is needed to determine the applicability of digital twin training for more complex tasks. Another limitation is our sample size and demographic. A larger and more diverse sample would provide a stronger argument for our experiment.\\
\indent
Our experiment was also limited by the mechanical differences of the virtual and physical rover. While great care was taken to model all movements as closely as possible, there were still some differences in the movement of the arm joints and gripper behavior. Observing the room was also slightly different between the rovers, with the physical rover having an imperfect pan-tilt system that only moved in 2 degree increments as opposed to the virtual rover's much smaller increments.\\
\indent
The largest limitation of our experiment comes from the experiment design itself. Our experiment shows that participants were able to improve performance in our task after going through a virtual practice run, when compared to participants that did not get a practice run at all. Further work is needed to study how impactful digital twin training is compared to real physical training.

}
\section{{Conclusions and Future Work}}
{
In this work, we investigated the use of digital twins for problem-solving and training in lunar surface telerobotics. Our robot, Armstrong, is controlled via a wireless gaming controller and VR headset. We conducted an experiment comparing task completion times between groups who practiced with our digital twin simulation and those who did not prior to using the physical rover. The results showed that practicing with the digital twin significantly reduced task completion time on the physical rover compared to the baseline. Digital twins offer a cost-effective solution for training future lunar rover operators, potentially reducing reliance on expensive facilities like JPL’s Mars Yard. Our findings have practical applications for upcoming missions, such as the FARSIDE lunar radio array, where 256 dipole antennas will be deployed semi-autonomously on the Moon’s far side. Simulations similar to Armstrong could help operators rehearse and troubleshoot, ultimately increasing mission success rates for future lunar telerobotics. 

Future work for this project includes applying the same digital twin technology to flight-ready rovers, such as Lunar Outpost's MAPP rover, in order to prepare operators for potential problems on the moon. This includes a higher level of accuracy in nearly every degree. Environmental characteristics of the lunar environment will be  tuned for Lunar Outpost's specific mission requirements, including shadows, celestial body positions overhead, dust dynamics, lens flare, gravity, and terrain features. This simulation, and its impacts on the performance of rover operators, is the subject of on-going work.

\section{{Acknowledgments}}
We thank Michael Walker for his thoughtful advice and encouragement during this experiment. This project was directly supported by the NASA Solar System Exploration Research Virtual Institute cooperative agreement 80ARC017M0006. We also received funding via the Lockheed Martin Corporation Space Division.

\newpage
\appendix
\section{Survey}
\label{sec:surveys}
\subsection{Virtual/Physical comparison}
\label{subsec:comparison}
\textit{Virtual/Physical Comparison, rated 1 as strongly disagree, 7 as strongly agree}
\begin{enumerate}
    \item Operating the virtual arm felt the same as operating the physical arm.
    \item Realigning the physical antenna felt the same as realigning the virtual antenna.
    \item Observing the room through the camera of the virtual rover felt the same as the physical rover.
    \item Driving the virtual rover felt the same as driving the physical rover.
\end{enumerate}

\subsection{Physical Only questions}
\label{subsec:phys_only}
\textit{Questions given \textit{only} to participants who only operated the physical rover. Rated 1 as strongly disagree, 7 as strongly agree.}
\begin{enumerate}
    \item It was easy to drive the rover.
    \item It was easy to operate the robotic arm.
    \item It was easy to grip the antenna.
    \item It was easy to rotate the antenna.
    \item It was easy to observe the room using the VR headset.
\end{enumerate}

\subsection{System Usability Scale (SUS) \citep{brooke1996sus}}
\label{subsec:sus}
\textit{Participants were asked to answer the following questions on a scale of 1 to 5, with 1 being strongly disagree, and 5 being strongly agree.}
\begin{enumerate}
    \item I think that I would like to use this system frequently.
    \item I found the system unnecessarily complex.
    \item I thought the system was easy to use.
    \item I think that I would need the support of a technical person to be able to use this system.
    \item I found the various functions in this system were well-integrated.
    \item I thought there was too much inconsistency in this system.
    \item I would imagine that most people would learn to use this system very quickly.
    \item I found the system very cumbersome to use.
    \item I felt very confident using the system.
    \item I needed to learn a lot of things before I could get going with this system.
\end{enumerate}

\subsection{Researcher Questions}
\label{subsec:researcher}
\textit{Questions generated by the research group to gauge the functionality of the rover. Rated 1 as strongly disagree, 7 as strongly agree.}
\begin{enumerate}
    \item The robot had all the capabilities needed to complete the task.
    \item The training I received was sufficient to complete the task.
    \item I understood the layout of the room.
\end{enumerate}

\subsection{Task Load Index (TLX) \citep{hart1988development}}
\label{subsec:TLX}
\textit{Please answer the following questions:}
\begin{enumerate}
    \item How mentally demanding was the task? (1 = Very Low, 7 = Very High)
    \item How physically demanding was the task? (e.g., eye strain, neck pain, etc.) (1 = Very Low, 7 = Very High)
    \item How hurried or rushed was the pace of the task? (1 = Very Low, 7 = Very High)
    \item How successful were you in accomplishing the task? (1 = Perfect, 7 = Failure)
    \item How hard did you have to work to accomplish your level of performance? (1 = Very Low, 7 = Very High)
    \item How insecure, discouraged, irritated, stressed, and annoyed were you? (1 = Very Low, 7 = Very High)
\end{enumerate}

\subsection{Demographics}
\label{subsec:demographics}
\textit{Please answer the following:}
\begin{itemize}
    \item What gender do you identify with? (Male, Female, Prefer not to say, Other)
    \item How old are you today? \_\_\_\_ (Example: 22)
    \item Are you a US citizen? (Yes/No)
    \item How often have you used VR? (1 = Never, 7 = Often)
    \item How often do you play video games with a controller? (1 = Never, 7 = Often)
\end{itemize}

\section{Affect Grid}
\label{sec:affect_grid}

The Affect Grid is a tool used to assess participants' emotional states during the experiment \citep{affectgridrussell}. It captures dimensions such as pleasure-displeasure and arousal-sleepiness on a 9-point scale. Participants mark one square to indicate how they are feeling at the present time. Below is the modified Affect Grid used in this study.

\begin{figure}[H]
    \centering
    \includegraphics[width=0.7\textwidth]{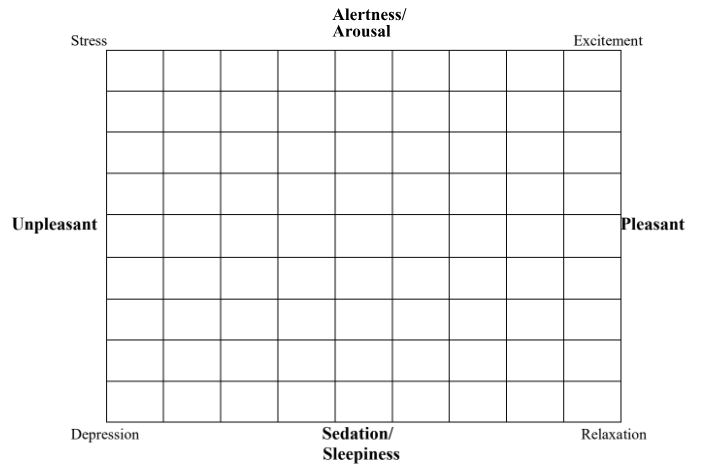} 
    \caption{Sample Affect Grid used for measuring emotional states during the experiment.}
    \label{fig:affect_grid}
\end{figure}

\section{Video}
\label{sec:video}
 Here is the link to our informational video detailing the operation of the rover, as well as the experimental task. Narrated by Katy McCutchan.

\url{https://www.youtube.com/watch?v=6YKsZBBNcwU&ab_channel=NetworkforExplorationandSpaceScience}

\bibliographystyle{jasr-model5-names}
\biboptions{authoryear}
\bibliography{refs}

\end{document}